\documentclass[10pt,journal,compsoc]{IEEEtran}

\ifCLASSOPTIONcompsoc
  \usepackage[nocompress]{cite}
\else
  \usepackage{cite}
\fi

\usepackage[pagebackref=true,breaklinks=true,bookmarks=false]{hyperref}
\usepackage{color}
\usepackage{amsmath}
\usepackage{array}

\usepackage{graphicx}
\usepackage{algorithm}
\usepackage{algorithmic}
\usepackage{booktabs}
\usepackage{multirow}
\usepackage{amsfonts}
\usepackage{amssymb}
\usepackage{extarrows}
\usepackage{nicefrac}

\newcommand{\E}{\mathbb{E}}
\newcommand{\D}{\mathbb{D}}

\newcommand{\YB}[1]{\textcolor{black}{#1}}

\makeatletter
\newcommand{\ssymbol}[1]{$^{\@fnsymbol{#1}}$}
\makeatother

\newcommand{\myparagraph}[1]{\textbf{#1}\hspace{1.8ex}}

\usepackage{threeparttable}
\usepackage{xcolor,colortbl}
\begin{document}
\title{
ZeroNLG: Aligning and Autoencoding Domains for Zero-Shot Multimodal and Multilingual Natural Language Generation}

%
% ~\IEEEmembership{Senior Member,~IEEE,}
\author{
% \textcolor{blue}{\url{https://github.com/yangbang18/ZeroNLG}}
% \\
Bang~Yang\ssymbol{1}, 
Fenglin~Liu\ssymbol{1}, 
Yuexian~Zou,
Xian~Wu,
Yaowei~Wang,
and~David~A.~Clifton
\IEEEcompsocitemizethanks{
\IEEEcompsocthanksitem 
\ssymbol{1} Equal Contributions. Ordered by a coin toss.
\IEEEcompsocthanksitem 
Bang Yang and Yuexian Zou are with ADSPLAB, School of ECE, Peking
University Shenzhen Graduate School, Shenzhen 518055, China, and also with Pengcheng Laboratory, Shenzhen 518052, China. 
E-mail: \{yangbang, zouyx\}@pku.edu.cn.
\IEEEcompsocthanksitem Fenglin Liu and David A. Clifton are with the Department of Engineering Science, University of Oxford, OX3 7DQ Oxford, U.K.. DAC is also with the Oxford-Suzhou Centre for Advanced Research, Suzhou, China. 
E-mail: \{fenglin.liu, david.clifton\}@eng.ox.ac.uk.
\IEEEcompsocthanksitem Xian Wu is with Tencent AI Lab, China. E-mail: kevinxwu@tencent.com.
\IEEEcompsocthanksitem Yaowei Wang is with Pengcheng Laboratory, Shenzhen 518052, China. 
E-mail: wangyw@pcl.ac.cn.}% <-this % stops an unwanted space
\thanks{Manuscript received XXX; revised XXX. (Corresponding author: Yuexian Zou.)}}

\markboth{Journal of \LaTeX\ Class Files,~Vol.~14, No.~8, August~2015}%
{Shell \MakeLowercase{\textit{et al.}}: Bare Demo of IEEEtran.cls for Computer Society Journals}
 
\IEEEtitleabstractindextext{%
\begin{abstract}
Natural Language Generation (NLG) accepts input data in the form of images, videos, or text and generates corresponding natural language text as output. Existing NLG methods mainly adopt a supervised approach and rely heavily on coupled data-to-text pairs. However, for many targeted scenarios and for non-English languages, sufficient quantities of labeled data are often not available. As a result, it is necessary to collect and label data-text pairs for training, which is both costly and time-consuming. To relax the dependency on labeled data of downstream tasks, we propose an intuitive and effective zero-shot learning framework, ZeroNLG, which can deal with multiple NLG tasks, including image-to-text (image captioning), video-to-text (video captioning), and text-to-text (neural machine translation), across English, Chinese, German, and French within a unified framework. ZeroNLG does not require any labeled downstream pairs for training. During training, ZeroNLG (i) projects different domains (across modalities and languages) to corresponding coordinates in a shared common latent space; (ii) bridges different domains by aligning their corresponding coordinates in this space; and (iii) builds an unsupervised multilingual auto-encoder to learn to generate text by reconstructing the input text given its coordinate in shared latent space. Consequently, during inference, based on the data-to-text pipeline, ZeroNLG can generate target sentences across different languages given the coordinate of input data in the common space. Within this unified framework, given visual (imaging or video) data as input, ZeroNLG can perform zero-shot visual captioning; given textual sentences as input, ZeroNLG can perform zero-shot machine translation. We present the results of extensive experiments on twelve NLG tasks, showing that, without using any labeled downstream pairs for training, ZeroNLG generates high-quality and ``believable'' outputs and significantly outperforms existing zero-shot methods. Our code and data are available at \url{https://github.com/yangbang18/ZeroNLG}.
\end{abstract}

}

\maketitle

\IEEEpeerreviewmaketitle

\IEEEraisesectionheading{\section{Introduction}\label{sec:introduction}}

\IEEEPARstart{N}{atural} language generation (NLG), also known as the data-to-text generation, aims to comprehend the content of provided data, which may come in various forms such as text \cite{bahdanau2015neural}, image \cite{Vinyals2015Show}, and video \cite{Venugopalan2015Translating}, and produce coherent text in natural language automatically. 
NLG has a wide range of applications, including image and video captioning and machine translation. 
Due to its broad usage scenarios, NLG has been receiving extensive research interests \cite{mokady2021clipcap,yang2021non,hu2022scaling,costa2022no}. 
Existing NLG approaches usually adopt an encoder-decoder framework\cite{bahdanau2015neural}, where the encoder calculates vector representations for the input data, and the decoder employs RNNs \cite{Hochreiter1997LSTM} or Transformers \cite{Vaswani2017transformer} to generate the target sentences using the encoded representation. 
Such approaches have demonstrated state-of-the-art performance in various natural language generation tasks \cite{hu2022scaling,stahlberg2020neural,chowdhery2022palm,flamingo}.

Most existing encoder-decoder-based approaches are purely data-driven and their performance is heavily reliant on the volume and quality of available labeled data-text pairs. However the acquisition of paired data can be time-consuming and costly in real-world situations, and its scarcity can prohibit the scale of models; with less restrictive approaches, we could dramatically increase the size of available data used to train, and thereby substantially enlarge model scalability to these increased dataset sizes.
Although numerous datasets of data-English pairs have been made available publicly, the availability of data-text pairs in non-English languages is often relatively uncommon or may even be unavailable. Consequently, the lack of labeled training data poses a significant challenge to developing NLG models for non-English languages - contributing to the highly uneven representation of less commonly-employed languages, which can in itself be a barrier to ``fair AI'' that would be usable by otherwise marginalized and under-represented communities.
For instance, to generate video captions in Chinese/French/German or to translate Chinese text into French/German, it is necessary to collect video-Chinese/French/German sentence pairs or Chinese-French/German sentence pairs, respectively.  This problem scales in difficulty when considering even less commonly-encountered languages, many of which correspond directly to less privileged communities.
To relax the dependence on labeled data for downstream NLG tasks, in this paper we propose the novel ZeroNLG framework, which is particularly useful for non-English languages where data-text pairs are limited in availability.

\begin{figure}[t]
\centering
\includegraphics[width=1\linewidth]{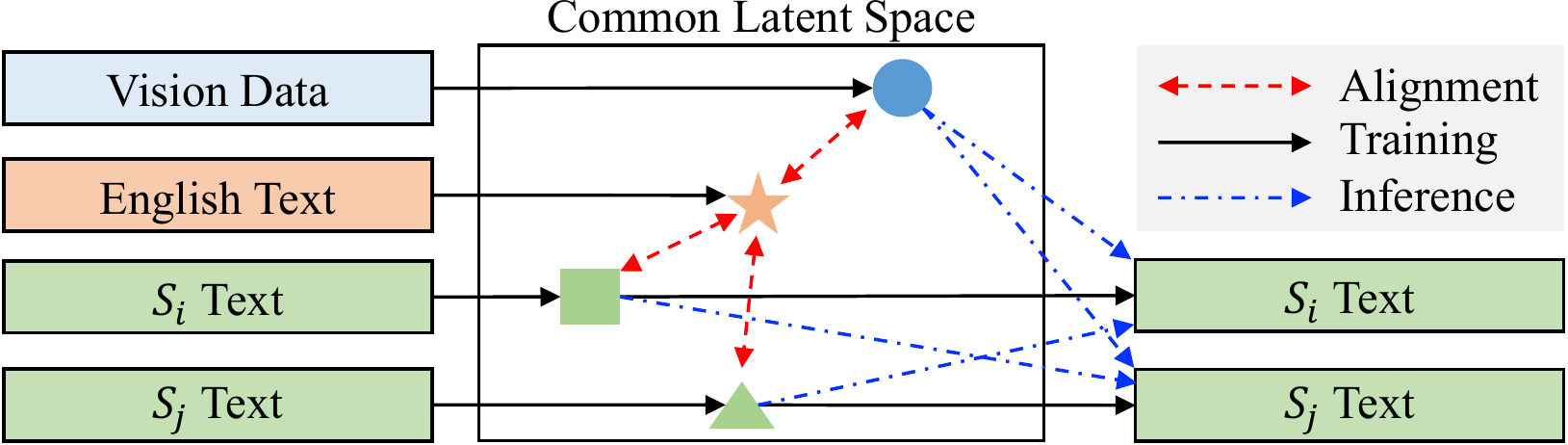}
\vspace{-15pt}
\caption{
During training, ZeroNLG first (i) projects different data across modalities and languages to corresponding coordinates in a shared common latent space; (ii) aligns their coordinates to bridge different domains; Here $S_i$ and $S_j$ refer to the text in non-English text, e.g. Chinese and Germany; (iii) performs unsupervised auto-encoding to learn to generate/reconstruct text given the coordinate of input text in this space. During inference, ZeroNLG encodes the input data acquiring its coordinate in this space, which can be directly used to perform zero-shot data-to-text generation (i.e., visual captioning and machine translation) without the need for downstream labeled pairs.}
\label{fig:introduction}
\end{figure}

To conduct zero-shot NLG across different modalities and languages, the core objective is to bridge the gap between various domains, e.g., vision and language domains, English and Chinese domains, and Chinese and German domains.
To this end, as shown in Figure~\ref{fig:introduction}, we propose the Cross-Domain Alignment pre-training objective,
which adopts Mean Square Error (MSE) and Info Noise Contrastive Estimation (InfoNCE) \cite{oord2018representation} losses, to pre-train a vision encoder and a multilingual encoder.
In this way, the aligned and bridged domains can be used for zero-shot multimodal and multilingual natural language generation.
In implementations, we introduce a vision encoder, an English encoder, and a multilingual encoder.
The motivation for introducing an English encoder comes from the fact that there are lots of English-centric corpus, e.g., image-to-English datasets \cite{chen2015microsoft,young2014flickr}, video-to-English datasets \cite{Xu2016MSR-VTT,wang2019vatex}, and non-English-to-English translation datasets \cite{Vaswani2017transformer,Ott2019fairseq,wu2019paylessattention}.
Therefore, we introduce the English encoder to make use of the existing English resources to pre-train the vision encoder and the multilingual encoder.
We first exploit the vision-English pairs ($D_1$) to pre-train the vision encoder and English encoder to \textit{align the vision and English domains}.
Then, we fix the parameters of the English encoder and exploit the English - non-English, e.g., the English-Chinese pairs ($D_2$), the English-German pairs ($D_3$), and the English-French pairs ($D_4$), 
to pre-train the multilingual encoder to \textit{align and bridge domains between English and any language}.
It is worth noting that the used datasets  $D_1$, $D_2$, $D_3$, and $D_4$ are independent and can have no overlap, i.e., English text in $D_1$, the English text in $D_2$, the English text in $D_3$, and the English text in $D_4$,  are separate sets and have no overlap. 
It means that there are no pairs of vision and Chinese/German/French; Chinese and German/French; German and French.
Considering that we have aligned the vision and English domains, our method can \textit{align and bridge the vision and non-English domains} without training on the pairs of vision and non-English text.
Meanwhile, due to various non-English domains being aligned with the English domain, \textit{the non-English domains, i.e., Chinese, German, and French, are aligned with each other}.
As a result, the ZeroNLG can \textit{align and bridge different domains across modalities and languages} in a shared common latent space without the training on downstream data-text pairs.

After aligning and bridging various domains, as shown in Figure~\ref{fig:introduction}, we further propose an unsupervised training objective \YB{Denoising} Language Reconstruction (\YB{DLR}) to learn to conduct zero-shot NLG.
Here we present a multilingual auto-encoder, including the pre-trained multilingual encoder and a randomly initialized multilingual decoder.
The \YB{DLR} aims to reconstruct the input sentences across various languages. 
During training, our method samples the English/Chinese/German/French sentences $S_1/S_2/S_3/S_4$ from $D_1/D_2/D_3/D_4$, as input to learn to reconstruct the input sentences in the $S_i \to S_i$ ($i=1,2,3,4$) auto-encoding pipeline.
In the prediction stage, due to we have aligned and bridged various domains, we can directly replace the $S_i$ with images/videos $V$ as input to generate the zero-shot visual captions for different languages in the $V \to S_i$ ($i=1,2,3,4$) pipeline.
Meanwhile, we can perform the zero-shot Chinese $\leftrightarrow$ German, Chinese $\leftrightarrow$ French, and German $\leftrightarrow$ French machine translation by inputting $S_j$ to the model to generate the translation in the $S_j \to S_i$ ($j \neq i$) pipeline.
Overall, the proposed ZeroNLG can perform zero-shot multimodal and multilingual natural language generation without the requirements of any downstream data-text pairs for training.
Besides, we can find that our method has the potential to be easily extended to other languages (e.g., Swedish and Italian) by aligning and bridging the English and target language domains. 
The extensive experiments on various NLG tasks, including image captioning, video captioning, and machine translation, across English, Chinese, German, and French, significantly prove the effectiveness of the proposed ZeroNLG.

Overall, the contributions of this work are:
\begin{itemize}
    \item We propose an effective method ZeroNLG to make the first attempt to perform zero-shot multimodal and multilingual natural language generation in a unified framework, where the downstream training pairs are not available.

    \item Our method bridges different domains across modalities and languages by aligning them in a common latent space; then learns to perform zero-shot language generation by auto-encoding/reconstructing the sentences in different languages.

    \item The extensive experiments and analyses on twelve natural language generation tasks across multiple languages show that our ZeroNLG can generate desirable sentences without using any labeled downstream data-text pairs for training and significantly outperforms existing state-of-the-art zero-shot learning methods. 
\end{itemize}

\section{Related Works}
We introduce the related works from two aspects: i) natural language generation and ii) zero-shot learning.
\vspace{-5pt}
\subsection{Natural Language Generation (NLG)}
The goal of NLG is to automatically generate fluent and accurate natural language text based on given input data such as text \cite{bahdanau2015neural}, images \cite{Vinyals2015Show}, and video \cite{Venugopalan2015Translating}.

This is typically achieved using an encoder-decoder framework where the encoder computes intermediate representations of the input data and the decoder uses RNN \cite{Hochreiter1997LSTM} to generate the final output. 
Attention mechanisms \cite{Luong2015Seq2Seq_Attention,liu2020Prophet,xu2015show,li2023unify} have been proposed to provide the decoder with full access to the source information, resulting in more efficient use of the input data. 
In particular, fully attentive models such as the Transformer \cite{ashish2017attention} have been successful in achieving state-of-the-art results in multiple NLG tasks such as image captioning \cite{Cornia2020M2,liu2019MIA}, video captioning \cite{Liu2021o2na,yang2022clip}, and neural machine translation \cite{wu2019paylessattention,Ott2019fairseq}. 
However, to efficiently train the data-driven models, most existing works rely on pairs of input data and corresponding output text, which could be difficult to obtain in the real world.
Additionally, there has been relatively little research concerning zero-shot learning.

\subsection{Zero-shot Learning}
Recently, few-shot learning \cite{Wang2020Fewshot} has received growing research interests \cite{Dhillon2020fewshot,Tian2020fewshot, Perez2021True,Gu2022PPT,Gao2021Making,Tsimpoukelli2021Multimodal}. Inspired by the success of few-shot learning, several works \cite{philip2020monolingual,chen2022towards} explored such an approach for data-to-text NLG tasks, which mainly include text-to-text and vision-to-text tasks.

In recent years, lots of zero-shot text-to-text machine translation models have been proposed \cite{johnson2017GoogleMT,Artetxe2018UnsupervisedNMT,Lample2018UnsupervisedNMT,Lample2018Phrase-Based,philip2020monolingual,chen2022towards}. Typically, the source language and the target language are mapped into a common latent space, where sentences with the same semantic meaning are well aligned, thus the zero-shot text-to-text translation can be carried out. 
Another line of research focuses on the ``prompt-based learning''\cite{liu2023pre} of large language models (LLMs) that acquire impressive sentence completion ability from massive pre-training text data. By providing LLMs with a textual template that consists of task-specific information \cite{raffel2020exploring,li2021prefix}, several in-context examples \cite{brown2020language,flamingo}, or a chain of thoughts \cite{Wei2022CoT,kojima2022CoT,chowdhery2022palm}, prompt-based methods can exploit the potentials of LLMs to perform zero-shot or few-shot NLG. 

For the vision-to-text task (i.e., visual captioning), zero-shot learning is particularly difficult because of the great disparities between the vision and the language domains, as well as the distinct characteristics of each modality.
As a result, the zero-shot vision-to-text works \cite{gu2018unpaired,Feng2019Unsupervised_ic,gu2019unpaired,liu2019exploring,nukrai2022text,tewel2022zerocap,su2022language,yu2022coca,liu2022aligning,zeng2023socratic} are relatively much less and the overall frameworks are more complex than those used for text-to-text tasks.
For example, \cite{gu2018unpaired} proposed a method to generate captions in a central language (Chinese) and subsequently translate them into the target language (English), without the training on downstream vision-text pairs.
However, this proposed method can not adopt visual information from images to generate more robust captions containing accurate visual details.
\cite{Feng2019Unsupervised_ic} aligned and bridged the vision and language domains with visual objects (e.g., \textit{girl}, \textit{umbrella}).
To achieve it, it was necessary to employ complex models and strategies to obtain higher-quality vision captions, e.g. object detection model \cite{huang2017speed}, image reconstruction \cite{Feng2019Unsupervised_ic} and adversarial learning \cite{goodfellow2014GAN}, in which the detector is limited to a pre-defined set of objects. 
Nevertheless, it does not incorporate other visual information, e.g., attributes (\textit{wooden}), relations (\textit{holding}), and color (\textit{red})) to include more visual details to generate captions.
\cite{gu2019unpaired,Yang2020USGAE} used the scene graph to bridge the gap between the vision and language domains. 
However, in order to construct an accurate and reliable scene graph, they were obliged to utilize Faster-RCNN \cite{ren2015faster} as the object detector, MOTIFS \cite{Zellers2018Motifs} as the relationship detector, and an additional classifier for attribute identification \cite{Yang2020USGAE}. 
\cite{Laina2019Towards,liu2022aligning} require the adversarial learning \cite{goodfellow2014GAN,Zhu2017unpairedI2I} for training.
Most recently, several works \cite{tewel2022zerocap,zeng2023socratic,su2022language} adopt LLMs like GPT \cite{radford2019language,brown2020language} for zero-shot visual captioning. 
While these techniques are effective, they have some drawbacks including the excessive parameterization of LLMs and a lack of adaptability to multilingual environments.
Overall, although the existing methods for zero-shot vision-to-text have shown considerable progress, they are hard to implement and still far from real-world applications.

To this end, we propose the ZeroNLG framework. 
The unique advantages of our method are i) it is simple but highly effective, outperforming all existing zero-shot methods; ii) it can utilize the full information of input data to perform zero-shot generation; iii) it can deal with various NLG problems across modalities and languages within a unified framework; iv) it could easily be extended to other languages - we prove its effectiveness on Chinese, German, and French.
As a result, our method could have the potential to promote the application of NLG, especially vision-to-text, for various low-resource language applications.

\begin{figure*}[t]
\centering
\includegraphics[width=1\linewidth]{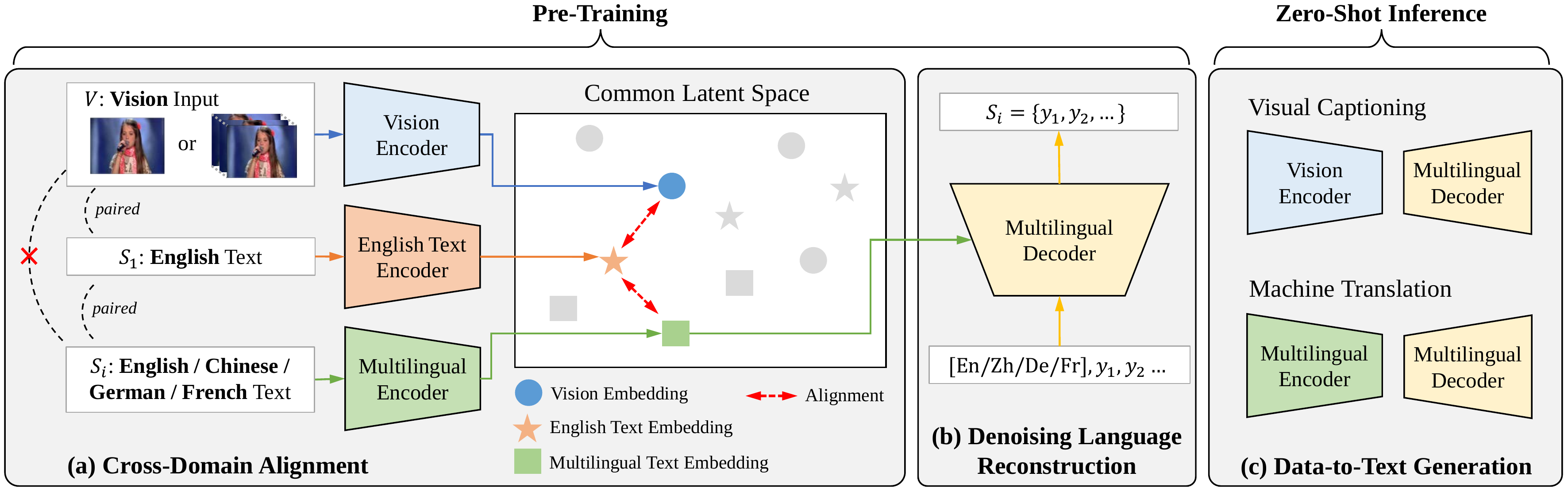}
\vspace{-15pt}
\caption{The illustration of our proposed ZeroNLG, including two components: cross-domain alignment and \YB{denoising} language reconstruction, where the former aims to align and bridge different data in a shared common latent space and the latter aims to reconstruct the input sentences across different languages, learning to generate sentences based on the embeddings in the common latent space.
We rely on English-centric pairs for training, i.e., vision-English, English-Chinese, English-German, and English-French, where the English texts in different sets have no overlap.
During inference, we can perform zero-shot natural language generation, including vision-to-Chinese/German/French captioning, and Chinese $\leftrightarrow$ German, Chinese $\leftrightarrow$ French, and German $\leftrightarrow$ French machine translation.
}
\label{fig:framework}
\end{figure*}

\section{Approach}
We first formulate how ZeroNLG tackles zero-shot NLG and then introduce two key components: cross-domain alignment and \YB{denoising} language reconstruction.

\subsection{Formulation}
As shown in Figure~\ref{fig:framework}, our ZeroNLG first aligns and bridges different domains in a shared common latent space, then performs unsupervised auto-encoding to learn to generate text by reconstructing the input text. In implementations, we choose English, Chinese, German, and French to evaluate our approach.
Therefore, we denote the vision data as $V$, the English sentence as $S_1$, the Chinese sentence as $S_2$, the German sentence as $S_3$, and the French sentence as $S_4$. 
As shown in Figure~\ref{fig:framework}, ZeroNLG includes four modules, i.e., a vision encoder $\E_v(\cdot)$, an English text encoder $\E_e(\cdot)$, a multilingual encoder $\E_m(\cdot)$, and a multilingual decoder $\D_m(\cdot)$.
Our ZeroNLG is defined as follows:
%%%%%%%%%%%%%%%%%%%%%%%%%%%%%%%%%%%%%%%%%%%
\begin{equation}
\footnotesize
\begin{aligned}
\label{eq:formulation}
\textbf{Pre-training}&
\left\{
\begin{array}{ll}
\text{Vision-English Alignment}& : \E_v(V) \xleftrightarrow{\text{Align}} \E_e(S_1) \\
\text{Cross-Lingual Alignment} & : \E_e(S_1) \xleftrightarrow{\text{Align}} \E_m(S_i) \\
\text{\YB{Denoising} Reconstruction} & : \E_m(S_i) \xrightarrow[\D_m(\cdot)]{\text{Reconstruct}} S_i 
\end{array}
\right. 
\\
\begin{array}{ll}
\textbf{Zero-shot} \\
\textbf{Inference}
\end{array}
&
\left\{
\begin{array}{ll}
\text{Zero-shot Vision-to-Text}& : \E_v(V) \xrightarrow[\D_m(\cdot)]{\text{Generate}} S_j \\
\text{Zero-shot Text-to-Text} & : \E_m(S_i) \xrightarrow[\D_m(\cdot)]{\text{Generate}} S_j 
\end{array}
\right. 
\end{aligned}
\end{equation}
%%%%%%%%%%%%%%%%%%%%%%%%%%%%%%%%%%%%%%%%%%%
where $i,j=1,2,3,4$, $j \neq i$, $S_j$ denotes the target sentences.
It is worth noting that introducing the English text encoder $\E_e(\cdot)$ has several merits: 
1) we can make full use of existing English-centric resources, e.g., the pairs of vision and English, and the pairs of non-English and English;
2) we can adopt the English-centric pre-trained model, e.g., CLIP \cite{radford2021learning}, as our basic model to boost the alignment of vision and English domains.
% directly skip a specific stage by exploiting an off-the-shelf model, e.g., building ZeroNLG upon the models like CLIP\cite{radford2021learning} to exclude stage ${\rm \bf I}$.
Such a continual learning scheme is a critical step towards sustainable AI \cite{budennyy2023eco2ai}. 
After pre-training, our ZeroNLG can directly perform zero-shot vision-to-text and text-to-text in the $\D_m(\E_v(V)) \rightarrow S_j$ and $\D_m(\E_m(S_i)) \rightarrow S_j$ pipelines, respectively.
Therefore, the ZeroNLG can deal with zero-shot NLG across modalities and languages within a single framework without the requirement of downstream labeled data-text pairs, which are not easy to acquire in the real world, especially for non-English languages.

\subsection{Cross-Domain Alignment}
\label{sec:cross_domain_alignment}

We introduce the Info Noise Contrastive Estimation (InfoNCE) \cite{oord2018representation} and Mean Square Error (MSE) losses to align and bridge different domains.
In particular, InfoNCE loss is a type of contrastive loss function used for self-supervised learning and has shown success in bridging the gap between the visual and textual modalities \cite{radford2021learning,jia2021scaling,yang2022unified};
MSE loss minimizes the distance between two different distributions.

Given a batch of $K$ English-centric training samples, including $K$ pairs of English text and Data, we denote the encoded English text and Data of $k^{\text{th}}$ training sample as ($s_k,d_k$), where $s = \E_e(S_{1})$.
We take the vision data and the text data in $i^{\text{th}}$ language as an example, if the Data is vision data, $d = \E_v(V)$; if the Data is multilingual text data in $i^{\text{th}}$ language, $d = \E_m(S_i)$.
Therefore, the InfoNCE loss can be formulated as follows:
%%%%%%%%%%%%%%%%%%%%%%%%%5
\begin{equation}
% \footnotesize
\begin{aligned} 
   \mathcal{L}^{s \rightarrow d}_{\text{InfoNCE}} &= 
    - \frac{1}{K} \sum_{k=1}^K \log{\frac{\exp(\langle s_k, d_k\rangle / \tau)}{\sum_{l=1}^{K} \exp(\langle s_k, d_l\rangle / \tau)}}    \\
    \mathcal{L}^{d \rightarrow s}_{\text{InfoNCE}} &= - \frac{1}{K} \sum_{k=1}^K \log{\frac{\exp(\langle d_k, s_k\rangle/ \tau)}{\sum_{l=1}^{K} \exp(\langle d_k, s_l\rangle/ \tau)}} \\
    \mathcal{L}_{\text{InfoNCE}} &= \frac{1}{2} \left(
     \mathcal{L}^{s \rightarrow d}_{\text{InfoNCE}}+ \mathcal{L}^{d \rightarrow s}_{\text{InfoNCE}}
     \right) ,
\end{aligned}
\end{equation}
%%%%%%%%%%%%%%%%%%%%%%%%%
where the $\langle \cdot, \cdot \rangle$ and $\tau$ denote the cosine similarity and a temperature hyper-parameter \cite{Chen2020CL}, respectively. 

The MSE loss can be defined as follows:
%%%%%%%%%%%%%%%%%%%%%%%%%%%%
\begin{equation}
   \mathcal{L}_{\text{MSE}} = \frac{1}{2K}\sum_{k=1}^K ||s_k - d_k||_2^2 ,
\end{equation}
%%%%%%%%%%%%%%%%%%%%%%%%%%%%
where $||\cdot||_2$ denotes L2-norm.

By combing the InfoNCE and MSE losses, we obtain the final training loss of the Cross-Domain Alignment (CDA):
\begin{equation}
   \mathcal{L}_{\rm CDA} = \lambda_1 \cdot \mathcal{L}_{\text{InfoNCE}} + \lambda_2 \cdot \mathcal{L}_{\text{MSE}} , 
\end{equation}
where $\lambda_1, \lambda_2 \in [0,1]$ are hyper-parameters that control the strength of each loss item. Through the above equation, our method can align and bridge different domains in a shared latent space, which provides a solid bias for zero-shot natural language generation.

\subsection{\YB{Denoising} Language Reconstruction}
\label{sec:language_reconstruction}
After aligning multimodal and multilingual domains, the next key step is to enable ZeroNLG to learn to generate multilingual text.
Here, we introduce an unsupervised objective dubbed \YB{Denoising} Language Reconstruction (\YB{DLR}) to train ZeroNLG. 
Specifically, we adopt Transformer \cite{ashish2017attention} as the decoder to reconstruct $S_i$ ($i=1,2,3,4$) in the $S_i \to S_i$ auto-encoding pipeline.
In implementations, we first randomly sample the sentences in $i^{\text{th}}$ language $S_i$, then adopt the multilingual encoder $\E_m(\cdot)$ to acquire the coordinates (i.e., embeddings) of input sentences in the latent space $\E_m(S_i)$, and finally adopt the multilingual decoder $\D_m(\cdot)$ to reconstruct the input sentences, defined as:
\begin{equation}
     S_i=\D_m(\E_m(S_i)) ,
\end{equation}
where $\E_m(S_i)$ denotes the coordinates of input text in the latent space. 
To train our method, we take the input sentence as the ground truth to be generated, i.e., $S_i = \{y_0, y_1,y_2,\dots,y_{|S_i|}\}$, where $y_0$ and $|S_i|$ denote the begin-of-sentence token and the number of tokens, respectively, and utilize the cross-entropy loss, which is widely used in natural language generation problems:
%%%%%%%%%%%%%%%%%5
\begin{equation}
\label{eq:DLR}
\mathcal{L}_{\rm \YB{DLR}} = - \sum_{l=1}^{|S_i|} \log p(y_l \mid y_{0: l-1}; \E_m(S_i)),
\end{equation}
%%%%%%%%%%%%%%%%%
where we implement $y_0$ as a language-specific token following \cite{tang2020multilingual,fan2021beyond} so that the decoder can be aware of which language to be generated.

\myparagraph{Data Corruption}
For vision-to-text, due to (i) the large variations of images and videos caused by different object attributes, occlusion, motion blur etc \cite{meng2022adavit};
(ii) the great disparities between the vision and the language domains \cite{liang2022mind},
we propose two data corruption strategies to further improve the performance and robustness of our ZeroNLG.

In implementations, we simultaneously consider the input and feature corruptions. 
For input corruption, we adopt the masking strategy as in BERT \cite{devlin2018bert} to randomly mask $r\%$ tokens of the input sentences $S_i$, obtaining the corrupted input sentences $S'_i$.
As a result, the \YB{DLR} process becomes:
%%%%%%%%%%%%5
\begin{equation}
     S_i=\D_m(\E_m(S'_i)) ,
\end{equation}
%%%%%%%%%%%%
For the feature corruption, we propose to add Gaussian noise $n \sim \mathcal{N}(0, \epsilon)$ into the text features $\E_m(S_i)$ (i.e., the coordinates) of input sentences $S_i$,  acquiring the corrupted features of input sentences $\E'_m(S_i) = \E_m(S_i) + n$. 
Therefore, the \YB{DLR} process is re-defined as follows:
%%%%%%%%%%%%5
\begin{equation}
     S_i=\D_m(\E'_m(S_i)) = \D_m(\E_m(S_i) + n) ,
\end{equation}
%%%%%%%%%%%%
Through data corruption, we can encourage the model to learn more robust latent representations, achieving strong performance on zero-shot natural language generation.

\section{Experiments}
In this section, we conduct experiments on multiple NLG tasks, i.e., vision-to-text image captioning and video captioning, and text-to-text neural machine translation.
% We first describe two public datasets for pre-training and several public datasets used for evaluation.
We first describe public datasets for pre-training and evaluation.
Then, we present the performance of our approach on zero-shot multimodal and multilingual natural language generation across modalities and languages.

\subsection{Experimental Setups}
\label{sec:experimental_setup}

\subsubsection{Datasets and Downstream Tasks}
\label{sec:datasets}
\smallskip\noindent\myparagraph{Pre-Training Datasets} 
\YB{The WebImageText\cite{radford2021learning,schuhmann2021laion} dataset (WIT) is used for vision-English alignment. WIT consists of 400 million image-English text pairs collected from the internet. For ease of experimentation, we directly use CLIP\cite{radford2021learning} pre-trained on WIT. Besides, the CC3M dataset\cite{sharma2018conceptual} is used for cross-lingual alignment and denoising language reconstruction.}
For CC3M consisting of 3.3M English (En) sentences, as we consider three non-English languages: Chinese (Zh), German (De), and French (Fr), we split the corpus into three non-overlapping splits, each of which contains 1.1M English sentences and is translated to the corresponding language via Google Translator\cite{Wu2016google}. 
As a result, we can acquire 1.1M En-Zh pairs, 1.1M En-De pairs, and 1.1M En-Fr pairs. 
Moreover, we randomly sample a subset of 1.1M English sentences from CC3M to ensure that the data (En, En-Zh, En-De, En-Fr) for pre-training is balanced.

It is worth noting that more strong and robust performance could be achieved by using large-scale human-annotated translation datasets.
In our experiments, we find that our approach can achieve state-of-the-art zero-shot results with the machine-translated sentences, thus we do not attempt to utilize more human-labeled high-quality datasets for pre-training.
Therefore, our model is not limited to the currently used pre-training 
data.

%%%%%%%%%%%%%%%%%%%%%%%%%%
\begin{table}[t]
\centering
% \tabfootnotesize
% \footnotesize
% \setlength{\tabcolsep}{1pt}
% \renewcommand\tabcolsep{4.5pt}
\footnotesize
    \caption{Pre-training and testing datasets used for experiments. 
    It is worth noting that the four English corpora used for pre-training are independent and can have no overlap.
    For English application scenarios, as the training data and testing data are from different domains and we also do not use any downstream data for training, we can still consider the evaluation on the English corpora as zero-shot NLG, as in the existing works \cite{tewel2022zerocap,su2022language,tewel2022zero,yu2022coca,zeng2023socratic}.
    As a result, we evaluate ZeroNLG on twelve tasks across modalities and languages.
    As there are no human-annotated datasets available for the video to German/French tasks, we report the qualitative results in Figure~\ref{fig:qualitative_examples}.
    }
    \label{tab:statistics}
    
\begin{tabular}{@{}cllll@{}}
\toprule
 
 & \bf Data-Text Pairs  &  \bf Pre-training 
&  \bf  Testing \\
\midrule

\multirow{5}{*}{\rotatebox{90}{\begin{tabular}[c]{@{}c@{}} \bf English \\ \bf Corpora \end{tabular}}} & Image-English & WIT \cite{radford2021learning,schuhmann2021laion} & MS-COCO \cite{chen2015microsoft}\\
& English-Chinese & CC3M-Zh \cite{sharma2018conceptual}&  En-Zh \cite{lan2017fluency}\\
& English-German & CC3M-De \cite{sharma2018conceptual}& WMT16\cite{WMT16} \\
& English-French & CC3M-Fr \cite{sharma2018conceptual}&WMT17\cite{WMT17} \\
& Video-English & - & MSR-VTT \cite{Xu2016MSR-VTT } \\

\midrule
\multirow{7}{*}{\rotatebox{90}{\begin{tabular}[c]{@{}c@{}} \bf Non-English \\ \bf Corpora \end{tabular}}} 
& Video-Chinese & - & VATEX-Zh \cite{wang2019vatex}  \\
& Image-Chinese & - & Flickr30k-Zh \cite{lan2017fluency} \\
& Image-German & - & Flickr30k-De \cite{elliott2016multi30k} \\
& Image-French & - & Flickr30k-Fr \cite{elliott2016multi30k}  \\
& Chinese-German & - & WMT16\cite{WMT16}  \\
& Chinese-French &-  & WMT17\cite{WMT17} \\
& German-French &  - & WMT17\cite{WMT17}\\
\bottomrule
\end{tabular}
\end{table}
%%%%%%%%%%%%%%%%%%%%%

 %%%%%%%%%%%%%%%%%%%%%%%%%%%%%%
\begin{table*}[t]
    \centering   
    \scriptsize
    % \small
    \setlength{\tabcolsep}{5.5pt}
    \caption{Performance of zero-shot vision-to-text visual captioning across three non-English Languages, i.e., Chinese, German, and French. B-4, R-L, and C are short for BLEU-4, ROUGE-L, and CIDEr, respectively. Higher is better in all columns. All previous works can not deal with zero-shot captioning for non-English languages. For comparison, we re-implement three state-of-the-art zero-shot English captioning models equipped with Google Translator and a state-of-the-art machine translation model NLLB-200.
    As we can see, our ZeroNLG can simultaneously generate desirable visual captioning across different languages in a single unified framework and achieves the best zero-shot results.}
    \begin{tabular}{l c ccc ccc ccc ccc }
    \toprule
    \multirow{3}{*}[-5pt]{\bf Methods} & \multirow{3}{*}[-5pt]{\bf Year} 
    &\multicolumn{3}{c}{\bf Video-to-Text} &\multicolumn{9}{c}{\bf Image-to-Text}
    \\ 
    \cmidrule(lr){3-5}
    \cmidrule(lr){6-14}
    & &\multicolumn{3}{c}{Chinese}
    &\multicolumn{3}{c}{Chinese}
    &\multicolumn{3}{c}{German}
    &\multicolumn{3}{c}{French}
    \\
    \cmidrule(lr){3-5}
    \cmidrule(lr){6-8}
    \cmidrule(lr){9-11}
    \cmidrule(lr){12-14}
    & &B-4 &R-L & C
    &B-4 &R-L &C 
    &B-4 &R-L &C
    &B-4 &R-L &C  \\

    \midrule
     CoCa \cite{yu2022coca} + Google Translator  
    & 2022 &1.4 &15.3 &4.3 
    &2.9 &19.0 &9.8
    &3.5 &20.3 &11.0
    &2.4 &15.1 &19.8\\

    CoCa \cite{yu2022coca} + NLLB-200 \cite{costa2022no}  
    &2022 
    &0.0 &11.6 &1.2
    &0.8 &13.2 &2.6
    &3.3 &20.5  &10.5
    &2.3 &15.3 &18.8\\
    
    CLIP-Re \cite{su2022language} + Google Translator   
    &2022
    &2.7 &20.8 &9.6
    &2.9 &23.7 &15.2
    &2.1 &21.6 &13.1
    &1.8 &15.5 &21.9
    \\

    CLIP-Re \cite{su2022language} + NLLB-200 \cite{costa2022no} &2022
    &0.8 &16.1 &2.8
    &1.1 &17.5 &4.6
    &2.2 &21.7 &12.8
    &1.9 &15.6 &21.4
    \\
     
    CapDec \cite{nukrai2022text} + Google Translator  
    &2022
    &2.9 &22.3 &5.1
    &4.7 &26.6 &13.9
    &5.4 &26.7 &16.9
    &2.5 &18.2 &23.2
    \\
    
    CapDec \cite{nukrai2022text} + NLLB-200 \cite{costa2022no} &2022 
    &0.9 &17.1 &1.9
    &1.8 &19.3 &4.7
    &5.2 &27.0 &16.9
    &2.5 &18.5 &23.6
    \\
    \cmidrule(lr){1-14}

    ZeroNLG & Ours
    &\bf 7.1&\bf 29.6&\bf 9.8
    &\bf 8.4&\bf 31.8&\bf 18.0
    &\bf 5.7&\bf 27.2&\bf 17.1
    &\bf 2.8&\bf 18.6&\bf 24.8
    \\

    \bottomrule
    \end{tabular}
    \label{tab:main_non_english_caption}
\end{table*}
%%%%%%%%%%%%%%%%%%%%%%%%%%%%%%

\smallskip\noindent\myparagraph{Downstream Tasks and Datasets} As shown in Table~\ref{tab:statistics}, we consider two types of natural language generation tasks: visual captioning and machine translation.
We focus on four languages, i.e., Chinese, German, French, and English.
For visual non-English captioning, we report performance on video-to-Chinese, image-to-Chinese, image-to-German and image-to-French directions using VATEX-Zh\cite{wang2019vatex}, Flicrk30k-Zh\cite{lan2017fluency}, Flickr30k-De\cite{elliott2016multi30k}, and Flickr30k-Fr\cite{elliott2016multi30k} datasets, respectively.
For visual English captioning, we utilize the widely-adopted MSR-VTT \cite{Xu2016MSR-VTT} and MS-COCO \cite{chen2015microsoft} datasets to measure the performance of video-to-English and image-to-English, respectively. 
For evaluating the performance of our method on machine translation, we only adopt humanly translated and annotated data. 
We obtain the translation pairs used for evaluation from the WMT16\cite{WMT16} and WMT17\cite{WMT17} machine translation competitions and English-Chinese dataset \cite{lan2017fluency}, including 
(a) English-Chinese pairs, (b) English-German pairs, (c) English-French pairs, (d) Chinese-German pairs, (e) Chinese-French pairs, and (f) German-French pairs.
For data preparation, we adopt the official splits to split the datasets and only adopt the testing set to evaluate the zero-shot performance of our approach.

\subsubsection{Metrics}
For visual captioning in English, we follow the common practice in the literature to report BLEU-4 \cite{papineni2002bleu}, METEOR \cite{banerjee2005meteor}, ROUGE-L \cite{lin2004rouge} and CIDEr \cite{vedantam2015cider}. We also include the SPICE metric \cite{anderson2016spice} for the image-to-English generation. For visual non-English captioning, METEOR and SPICE metrics are excluded because they consider synonym matching and named entity recognition in English by default. All metrics are computed by the widely-used Microsoft COCO Evaluation Server \cite{chen2015microsoft}, where we use different toolkits to segment Chinese, German, and French sentences (as introduced next). For machine translation, we report the widely adopted BLEU \cite{papineni2002bleu} \YB{measured by the SacreBLEU toolkit\cite{post2018SacreBLEU}}.

\subsubsection{Implementation Details}
\label{sec:implementation_details}
As shown in Eq.~\ref{eq:formulation}, ZeroNLG includes a vision encoder, an English text encoder, a multilingual encoder, and a multilingual decoder. We implement them as follows.
\begin{itemize}
    \item Vision Encoder $\E_v(\cdot)$ and English text encoder $\E_e(\cdot)$: following existing works \cite{radford2021learning,nukrai2022text,su2022language,tewel2022zerocap,zeng2023socratic}, we adopt the pre-trained \YB{and frozen} CLIP \cite{radford2021learning}, which is composed of a ViT-B/32 model \cite{dosovitskiy2021an} and a 
    \YB{decoder-only text encoder\cite{radford2018improving},}
    to implement our vision encoder and English text encoder.
    Such practice can substantially save computing resources and energy, which is a critical step towards sustainable AI \cite{budennyy2023eco2ai}. 
    
    \item Multilingual encoder $\E_m(\cdot)$: \YB{we adopt the pre-trained multilingual DistilBERT\cite{sanh2019distilbert} as the multilingual encoder, which adopts WordPiece embeddings\cite{Wu2016google} and has a vocabulary of size 119,547. }

    \item Multilingual decoder $\D_m(\cdot)$: \YB{we implement it as Transformer decoder\cite{ashish2017attention} with 768 model dimensions, 12 attention heads, 3 layers, and the same word embeddings as that of the multilingual encoder.}

\end{itemize}

\YB{For the English text encoder, we extract global features from the position of the end-of-sentence token following CLIP. For the multilingual encoder, we truncate sentences into a maximum length of 128 and follow sBERT\cite{reimers2019Sentence-BERT} to obtain mean pooled global features.}
We use AdamW \cite{loshchilov2019decoupled} with L2 weight decay of 0.01 to train models for 3 epochs. 
We set the learning rate fixed to 2e-5 after 5K warm-up iterations. The batch size is 128 for cross-domain alignment and 32 for language reconstruction.
Based on the validation performance, we set $r=0$ and $\epsilon = 0.1$ for visual captioning and $r=5$ and $\epsilon = 0.01$ for machine translation. 
\YB{$\{\lambda_1, \lambda_2\}=\{1,0\}$ (Eq.~\ref{eq:formulation}) is used in CLIP for vision-English alignment. For cross-lingual alignment, we set $\{\lambda_1,\lambda_2\}=\{0,1\}$ (see Table~\ref{tab:more_ablation}).}
We use the Jieba toolkit\footnote{\url{https://github.com/fxsjy/jieba}} to segment Chinese sentences, and use the CoreNLP toolkit \cite{manning2014stanford} for German and French sentences. 
When processing videos, we uniformly sample 8 frames.
We use beam search with a beam size of 3 to generate texts. 
\YB{Our ZeroNLG can be trained} on \textit{only} an NVIDIA T4 card within 18 hours.

%%%%%%%%%%%%%%%%%%%%%%%%%
%%%%%%%%%%%%%%%%%%%%%%%%%
\begin{table*}[t]
    \centering  % 表居中
    \scriptsize
    \caption{Performance of zero-shot vision-to-text visual captioning in English. 
    \YB{$^\ddagger$: Our re-implementations}. 
    \ssymbol{1}: Pre-trained on LAION-2B \cite{schuhmann2022laionb} that consists of 2 billion image-English text pairs. 
    $^\dagger$:  SMs \cite{zeng2023socratic} needs to call GPT-3 API which is quite time-consuming. Therefore SMs \cite{zeng2023socratic} only reported their performance on 100 randomly sampled test instances. To conduct a fair comparison, we report the result of ZeroNLG on the same 100 test instances used by SMs.
    As we can see, our approach outperforms previous state-of-the-art zero-shot methods on most metrics.
    }
    \begin{tabular}{l c l ccccc cccc}
    \toprule
      \multirow{2}{*}[-3pt]{\bf Methods}  & \multirow{2}{*}[-3pt]{\bf Year}  
    &\multirow{2}{*}[-3pt]{\bf Pre-trained Backbone}
    &\multicolumn{4}{c}{\bf Video-to-Text (English)} &\multicolumn{5}{c}{\bf Image-to-Text (English)}
    
    \\
    % \cmidrule{3-17}
    \cmidrule(lr){4-7}
    \cmidrule(lr){8-12}
    & &  &B-4& M&R-L&C  &B-4&M&R-L&C&S
   \\
    % \hline
    \midrule

    CapDec\YB{$^\ddagger$} \cite{nukrai2022text} & 2022
    & CLIP
    &7.3 &14.1 &33.5 &8.4
    &8.8 &13.5 &33.0 &25.0 &7.9\\

    ZeroCap\YB{$^\ddagger$} \cite{tewel2022zerocap} & 2022
    &CLIP + GPT-2  
    &\YB{2.3}   &\YB{10.8}  &\YB{23.1}   &\YB{7.3}
    &\YB{1.8}   &\YB{9.1}   &\YB{19.7}   &\YB{14.4}   &\YB{5.0}
        \\

    MAGIC
    % $*$ 
    \cite{su2022language} & 2022
    &CLIP + GPT-2 
    &5.5    &13.3   & \bf 35.4   &7.4 
    &5.2   &12.5   &30.7   &18.3   &5.7
    \\

    EPT \cite{tewel2022zero} & 2022
    &CLIP + GPT-2 
    &3.0    &14.6   &27.7   & \bf 11.3
    &-&-&-&-&-
     \\

    CoCa\YB{$^\ddagger$} \cite{yu2022coca} & 2022 & CLIP-like\ssymbol{1}
    &3.4 &10.2 &21.5 &5.6
    &5.2 &11.2 &24.2 &16.7 &6.8
    \\  \cmidrule(lr){1-12}

    ZeroNLG & Ours & CLIP 
    &\bf 8.7 &\bf 15.0 &\bf 35.4 &9.9
    &\bf 9.6 &\bf 14.4 &\bf 34.9 &\bf 29.9 &\bf 8.7
    \\ \midrule [\heavyrulewidth]

     SMs$^\dagger$ \cite{zeng2023socratic} & 2023  & CLIP + GPT-3
    &-      &-      &-      &-
    &10.0   &\bf 16.2   &36.1   &50.1   &10.8\\ \cmidrule(lr){1-12}

     ZeroNLG$^\dagger$ & Ours  & CLIP   
    &-      &-  &-  &- 
    &\bf 12.0 &15.7 &\bf 38.1 &\bf 51.6 &\bf 11.1\\

    \bottomrule
    % \hline
    \end{tabular}
    \label{tab:main_english_caption}
\end{table*}
%%%%%%%%%%%%%%%%%%%%%%%%%
%$5%%%%%%%%%%%%%%%%%%%%%%%%$
\begin{table*}[t]
    \centering   
    \scriptsize
    % \small
    \setlength{\tabcolsep}{4pt}
    \caption{Performance of zero-shot machine translation across English (En), Chinese (Zh), German (De), and French (Fr). \YB{We report the BLEU metric measured by the SacreBLEU toolkit\cite{post2018SacreBLEU}}. 
    Higher is better in all columns.
    $\rightarrow$ and $\leftarrow$ denote the translation direction. 
    ($\cdot$) is calculated by comparing with our method in terms of the sentence pairs used for pre-training. 
    Our approach achieves encouraging performance with less number of parameters and pre-training pairs. 
    More importantly, these listed works can not deal with vision-to-text multimodal NLG tasks.
    }
    \begin{tabular}{l c c l ccc cc cc cc cc cc c}
    \toprule
    \multirow{2}{*}[-3pt]{\bf Methods}
    &\multirow{2}{*}[-3pt]{\bf Year}
    &\multirow{2}{*}[-3pt]{\bf \#Params}
    &\multirow{2}{*}[-3pt]{\bf \#Sentence Pairs}
    &\multicolumn{2}{c}{\bf En-Zh}
    &\multicolumn{2}{c}{\bf En-De}
    &\multicolumn{2}{c}{\bf En-Fr}
    & \bf Average 
    &\multicolumn{2}{c}{\bf Zh-De}
    &\multicolumn{2}{c}{\bf Zh-Fr}
    &\multicolumn{2}{c}{\bf De-Fr}
    & \bf Average \\
    \cmidrule(lr){5-6}
    \cmidrule(lr){7-8}
    \cmidrule(lr){9-10}
    \cmidrule(lr){11-11}
    \cmidrule(lr){12-13}
    \cmidrule(lr){14-15}
    \cmidrule(lr){16-17}
    \cmidrule(lr){18-18}
    &&&
    &$\rightarrow$ &$\leftarrow$
    &$\rightarrow$ &$\leftarrow$
    &$\rightarrow$ &$\leftarrow$
    & English
    &$\rightarrow$ &$\leftarrow$
    &$\rightarrow$ &$\leftarrow$
    &$\rightarrow$ &$\leftarrow$
    & Non-English \\
    \midrule
    
    mBART-50\cite{tang2020multilingual} &2020 &610.9M &203.7M (62x)
    &\bf 18.9 &\bf 12.5 
    &32.4 &34.0 
    &30.4 &41.1 
    &28.2
    &6.9 &0.3 
    &4.2 &1.7 
    &7.6 &17.9 
    &6.4
    \\
    
    M2M-100 \cite{fan2021beyond} & 2021 &483.9M &7.5B (2,000x)
    &16.4 &10.5 
    &24.5 &30.2
    &30.7 &36.4 
    &24.8
    &8.5 &\bf 13.3 
    &\bf 6.8 &14.9 
    &22.6 &23.5 
    &\bf 14.9\\
    
    NLLB-200\cite{costa2022no} &2022 & 617.2M &18B  (5,000x)
    &6.3 &12.8 
    &\bf 37.5 &\bf 39.8 
    &\bf 49.8 &\bf 46.8 
    &\bf 32.2
    &\bf 10.7 &4.1
    &5.7 &4.9
    &\bf 34.2 &\bf 30.8
    &14.7\\ 

    \cmidrule(lr){1-18}

    ZeroNLG  & Ours
    &\bf 165.0M & \bf 3.3M (1x) 
    &14.7&8.8 
    &20.5&21.1
    &22.0&24.6
    &18.6 
    &7.3 &11.9
    &5.2&\bf 16.2
    &16.7&18.5
    &12.6\\

    \bottomrule
    \end{tabular}
    \label{tab:main_translation}
\end{table*}
%$5%%%%%%%%%%%%%%%%%%%%%%%%$

\subsection{Zero-shot Vision-to-Text (Visual Captioning)}
\myparagraph{Vision-to-Non-English}
We conduct the experiments on three non-English languages, i.e., Chinese (Zh), German (De), and French (Fr). 
In detail, we adopt the VATEX-Zh \cite{wang2019vatex} to report the performance on video-to-Chinese; adopt the Flickr30k-Zh \cite{lan2017fluency}, Flickr30k-De \cite{elliott2016multi30k}, and Flickr30k-Fr \cite{elliott2016multi30k} to report the performance on image-to-Chinese, image-to-German, image-to-French, respectively.

Since all previous works can not deal with zero-shot visual captioning tasks for non-English languages, we re-implement three existing state-of-the-art zero-shot English captioning methods, i.e., CoCa \cite{yu2022coca}, CLIP-Re \cite{su2022language}, CapDec \cite{nukrai2022text}, and equip them with strong machine translation models to generate non-English captions.
In implementations, we use CoCa (ViT-B/32 variant) pre-trained on LAION-2B\cite{schuhmann2022laionb} and released by OpenCLIP \cite{ilharco2021OpenCLIP}, and re-implement the CLIP-Re and CapDec on the same corpus as ours. 
\YB{For translation,} we adopt a commercial translation product Google Translator \cite{Wu2016google} and a recent state-of-the-art machine translation model NLLB-200 \cite{costa2022no}.

Table~\ref{tab:main_non_english_caption} reports the zero-shot visual captioning results across Chinese, German, and French.
As we can see, 
our approach achieves the best zero-shot results across all metrics and languages.
For video captioning, our proposed ZeroNLG consistently outperforms the existing zero-shot captioning methods.
In detail, it achieves up to 4.2\%, 7.3\%, and 0.2\% absolute improvements compared to previous best results in terms of BLEU-4, ROUGE-L, and CIDEr, respectively. 
For image captioning, our ZeroNLG successfully surpasses all baselines across all languages by up to 18.4\%, 1.2\%, and 5.1\% relative improvements on the CIDEr metric in terms of Chinese, German, and French, respectively. 
In brief, the above results significantly prove the effectiveness of our approach ZeroNLG in dealing with zero-shot multimodal and multilingual natural language generation within a unified framework.

\smallskip\noindent\myparagraph{Vision-to-English}
Considering that several zero-shot English captioning methods have been proposed \cite{tewel2022zerocap,su2022language,tewel2022zero,yu2022coca,zeng2023socratic}, for comparison with these existing works, we further conduct the experiments on zero-shot vision-to-English to verify the effectiveness of our approach.
In particular, for a fair comparison, we follow previous works to conduct experiments on the MSR-VTT video captioning \cite{Xu2016MSR-VTT} and MSCOCO image English captioning \cite{chen2015microsoft} datasets.
It is worth noting that the training data and testing data are from different domains, therefore, like previous works \cite{tewel2022zerocap,su2022language,tewel2022zero,yu2022coca,zeng2023socratic}, we can consider this evaluation as zero-shot English captioning.
 
As shown in Table~\ref{tab:main_english_caption}, our method ZeroNLG substantially outperforms existing state-of-the-art zero-shot methods.
For example, on image captioning, our approach can outperform all previous works, several of which incorporate a large language modeling model -- GPT \cite{radford2019language,brown2020language}. On video captioning, although EPT\cite{tewel2022zero} achieves a better CIDEr score than ours (11.3 vs. 9.9), it needs more time to generate a caption for each video ($> 1$ minute in EPT vs. $< 1$ second in ZeroNLG). 
The vision-to-English experiments further prove the effectiveness of our approach, which achieves state-of-the-art zero-shot results across different languages.

%%%%%%%%%%%%%%%%%%%%%%%%%%%%%%%%%
\begin{figure*}[t]
\centering
\includegraphics[width=1.0\linewidth]{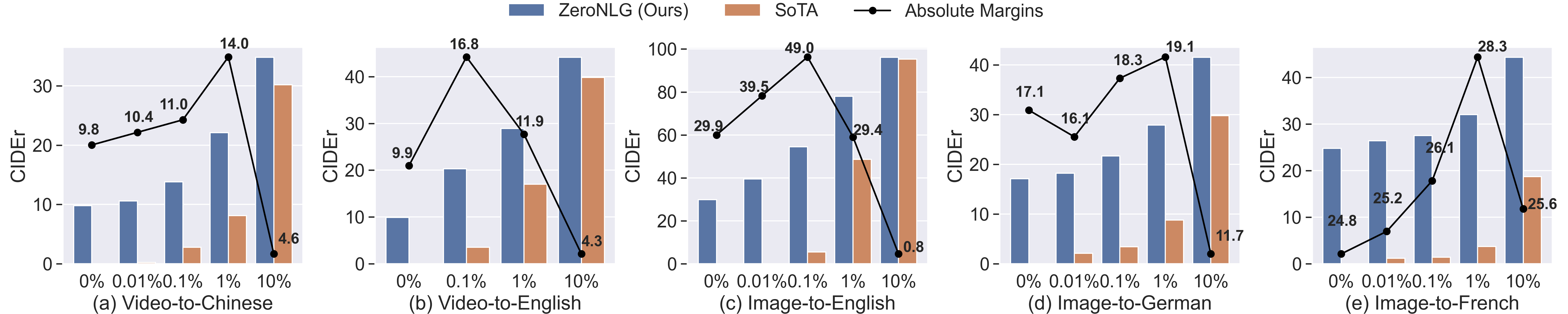}
\vspace{-15pt}
\caption{Results of vision-to-text visual captioning with respect to different ratios of downstream data used for training. The absolute margins between our model and the state-of-the-art (SoTA) model ClipCap\cite{mokady2021clipcap} are shown with the polyline.
Our method consistently and significantly outperforms the SoTA under the very limited pairs setting (i.e., 0.01\%, 0.1\%, and 1\%).
}
\label{fig:ablation_semi}
\end{figure*}
%%%%%%%%%%%%%%%%%%%%%%%%%%%%%%%%%
\begin{table*}[t]
    \centering   
    \scriptsize
    % \small
    \setlength{\tabcolsep}{2.5pt}
    \caption{
    Quantitative analysis of our ZeroNLG, which consists of the cross-domain alignment (CDA) and \YB{denoising} language reconstruction (\YB{DLR}). 
    }
    \begin{tabular}{l c c cccc ccccc cccc ccc ccc ccc ccc}
    \toprule
    \multirow{3}{*}[-5pt]{\bf Setting} 
    &\multirow{3}{*}[-5pt]{\bf CDA}
    &\multicolumn{5}{c}{\bf \YB{DLR}}
    &\multicolumn{7}{c}{\bf Video-to-Text}  &\multicolumn{14}{c}{\bf Image-to-Text}
    \\ 
    \cmidrule(lr){3-7}
    \cmidrule(lr){8-14}
    \cmidrule(lr){15-28}
    &
    &\multirow{2}{*}[-3pt]{\begin{tabular}[c]{@{}c@{}} \bf Data  \\ \bf Corruption \end{tabular}}
    &\multicolumn{4}{c}{\bf Languages}
    &\multicolumn{4}{c}{\bf English (En)} & \multicolumn{3}{c}{\bf Chinese (Zh)} & \multicolumn{5}{c}{\bf English (En)} & \multicolumn{3}{c}{\bf Chinese (Zh)} & \multicolumn{3}{c}{\bf German (De)} & \multicolumn{3}{c}{\bf French (Fr)} 

    \\ 
    \cmidrule(lr){4-7}
    \cmidrule(lr){8-11} 
    \cmidrule(lr){12-14}
    \cmidrule(lr){15-19}
    \cmidrule(lr){20-22}
    \cmidrule(lr){23-25}
    \cmidrule(lr){26-28}
    
    &&    
    &En &Zh &De &Fr
    &B-4&M&R-L&C & B-4 & R-L & C
    & B-4  &M & R-L & C & S
    & B-4  & R-L & C 
    & B-4  & R-L & C 
    & B-4  & R-L & C 
    
    \\ \midrule

    \bf Full &$\surd$  &$\surd$ &$\surd$ &$\surd$ &$\surd$ &$\surd$ 
    &\bf 8.7&\bf15.0&\bf35.4&\bf9.9
    &\bf7.1  &\bf 29.6  & \bf 9.8
    &\bf 9.6&\bf 14.4&\bf 34.9&\bf 29.9&\bf 8.7
    &\bf 8.4&\bf 31.8&\bf 18.0
    &\bf 5.7&\bf 27.2&\bf 17.1&
    \bf 2.8&\bf 18.6&\bf 24.8
    \\
    
    \midrule
    
    (a) &$\surd$ &- &$\surd$ &$\surd$ &$\surd$ &$\surd$
    &1.5&10.2&24.8&3.7
    &7.9 &18.6 &2.9
    &1.0&7.7&19.0&5.7&3.0
    &0.7&17.7&3.5
    &0.0&15.2&3.5
    &0.5&10.1&8.0
    \\ 
    
    (b) & -  &$\surd$ &$\surd$ &$\surd$ &$\surd$ &$\surd$ 
    &0.6&7.5&22.7&0.5
    &0.0  &16.8  &0.5
    &0.7 &5.3&15.8&1.0&1.0
    &0.0&14.9&0.7
    &0.0&11.4&0.7
    &0.0&7.2&2.8
    \\

    (c)  & -  & -  &$\surd$ &$\surd$ &$\surd$ &$\surd$  &\multicolumn{21}{c}{\bf Fail}
    \\

    \midrule
     
    (d) &$\surd$  &$\surd$ &$\surd$ & & &
    &7.3&14.6&34.0&9.0
    & -  & -  & -
    &9.3&14.2&34.3&27.5&8.3
    &-&-&-
    &-&-&-
    &-&-&-

    \\
    
    (e) &$\surd$  &$\surd$ & &$\surd$ & &
    &-&-&-&-
    &6.4 &28.1 &8.6
    &-&-&-&-&-
    &8.0&30.4&17.5
    &-&-&-
    &-&-&-
    \\

    (f) &$\surd$ &$\surd$ & & &$\surd$ &
    &-&-&-&-&-
    & -  & -  & -
    &-&-&-&-
    &-&-&-
    &3.6&25.3&13.8
    &-&-&-
    \\

    (g) &$\surd$  &$\surd$ & & & &$\surd$
    &-&-&-&-&-
    & -  & -  & -
    &-&-&-&-
    &-&-&-
    &-&-&-
    &2.3&17.8&21.2
    \\
    
    \bottomrule
    
    \end{tabular}
    
    \label{tab:ablation_study}
\end{table*}
%%%%%%%%%%%%%%%%%%%%%%%%%%%%%%%%%

\subsection{Zero-shot Text-to-Text (Machine Translation)}

To further prove the effectiveness of our approach, we conduct six zero-shot machine translation tasks across English (En), Chinese (Zh), German (De), and French (Fr), i.e., En-Zh, En-De, En-Fr, Zh-De, Zh-Fr, De-Fr.
We report the results on Table~\ref{tab:main_translation}. 
We compare our ZeroNLG with three pre-trained large language models (LLMs) designed for neural machine translation: mBART-50\cite{tang2020multilingual}, M2M-100\cite{fan2021beyond}, and NLLB-200\cite{costa2022no}. In detail, M2M-100 and NLLB-200 are respectively pre-trained on around 7.5B and 18B many-to-many training pairs, including En-Zh, En-De, En-Fr, Zh-De, Zh-Fr, and De-Fr, while mBART-50 and our ZeroNLG are only pre-trained on English-centric training pairs, i.e., En-Zh, En-De, and En-Fr, without any Zh-De, Zh-Fr, De-Fr training pairs.
Nevertheless, due to the training set and testing set being from different domains, we can consider all these LLMs and our ZeroNLG as zero-shot machine translation models.

As shown in Table~\ref{tab:main_translation}, our approach \YB{shows potential when compared} with the existing state-of-the-art LLMs, which adopt more model parameters and training pairs.
Especially, without the downstream pairs for training, our ZeroNLG can significantly outperform the mBART-50 model, across all non-English translation tasks. 
For the Zh$\leftarrow$Fr task, our ZeroNLG achieves the best results.
It further proves the effectiveness of our method in dealing with zero-shot machine translation, where the downstream pairs are not available.

\smallskip\noindent\myparagraph{Overall}
Combining the results of zero-shot vision-to-text and text-to-text results, we can conclude that our proposed ZeroNLG can perform zero-shot multimodal and multilingual natural language generation in a single framework and outperform previous state-of-the-art zero-shot methods.
The advantages under the scenarios without any downstream labeled data-text pairs show that ZeroNLG might be applied to other low-resource languages (Swedish, Italian, etc.).

\section{Analysis}
To understand the effect of each component in our framework, we conduct several analyses in this section.
We focus on the more challenging task, i.e., vision-to-text visual captioning, to perform the analysis.

\subsection{Semi-Supervised Learning}
\label{sec:semi}

To further prove the effectiveness of our method, we propose to utilize a few labeled downstream data-text pairs for fine-tuning.
To this end, in Figure~\ref{fig:ablation_semi}, we evaluate the performance of our approach on visual captioning across English, Chinese, German, and French with respect to the increasing amount of paired data. 
For a fair comparison, we also re-train the state-of-the-art (SoTA) model ClipCap \cite{mokady2021clipcap} using the same amount of downstream data-text pairs and the same architecture as our model.
As we can see, our ZeroNLG can be significantly boosted with few downstream labeled pairs and outperforms the SoTA under all ratios of data used for training. 
More importantly, i) without any downstream pairs for training (0\%), our ZeroNLG can even significantly surpass the SoTA trained with 10\% downstream pairs by 6.1\% absolute CIDEr score on image-to-French;
ii) under very limited pairs settings, e.g., 0.01\% and 0.1\%, our approach can outperform the SoTA trained with 1\% downstream pairs on all vision-to-text tasks.
The strong performance of our ZeroNLG under the very limited pairs setting proves the effectiveness of our approach in relaxing the reliance on the downstream pairs to provide a solid basis for natural language generation, which is particularly useful for low-resource language applications, where the labeled data-text pairs are scarce and hard to collect.

Overall, with very limited labeled downstream training pairs, our method can be efficiently fine-tuned and deployed on low-resource language applications.
Such capacity could improve the practicality of natural language generation in real-world applications, and contribute to marginalized and under-represented communities.

%%%%%%%%%%%%%%%%%%%%%%%%%%%%%
\begin{table*}[t]
    \centering
    \setlength{\tabcolsep}{3pt}
    \caption{\YB{Impact of encoder choices, the number of decoder layers ($L$), the corpora for cross-lingual alignment (CLA), and loss designs ($\lambda_1$, $\lambda_2$) of CLA. We perform the analysis on the image-to-text visual captioning under different languages and report the CIDEr scores. Default settings are highlighted in a gray background. As we can see, our ZeroNLG can benefit from more advanced pre-trained multilingual encoders, decoders with larger capacities, and proper pre-training corpora.
    }}
    \label{tab:more_ablation}
    \begin{minipage}{0.27\linewidth}{\begin{center}
        \begin{tabular}{l cccc}
        \toprule
        Encoder&En &Zh &De &Fr \\
        \midrule
        \rowcolor{gray!10}
        mDistilBERT &29.9 &18.0 &17.1 &24.8\\
        mBERT &32.0 &19.4 &\bf 18.0 &26.9\\
        XLM-R &28.7 &18.5 &16.5 &23.9\\
        sBERT$^*$ &\bf 32.7 &\bf 20.1 &\bf 18.0 &\bf 27.1 \\
        \bottomrule
        \multicolumn{5}{l}{\scriptsize \emph{$^*$: paraphrase-multilingual-mpnet-base-v2}}\\
        \\
        \end{tabular}
    \end{center}}
    \end{minipage}
    \begin{minipage}{0.2\linewidth}{\begin{center}
        \begin{tabular}{ccccc}
        \toprule
        $L$&En &Zh &De &Fr \\
        \midrule
        \rowcolor{gray!10} 
        3 &29.9 &18.0 &17.1 &24.8\\
        6 &31.0 &19.0 &17.9 &26.2\\
        12 &32.1 &19.2 &19.3 &28.8\\
        24 &\bf34.4 &\bf19.8 &\bf21.2 &\bf30.9\\
        \bottomrule
        \\
        \\
        \end{tabular}
    \end{center}}
    \end{minipage}
    \begin{minipage}{0.27\linewidth}{\begin{center}
        \begin{tabular}{lcccc}
        \toprule
        Corpora&En &Zh &De &Fr\\
        \midrule
        \rowcolor{gray!10} 
        (1) CC3M$_{\rm 4L}$ &29.9 &18.0 &\bf 17.1 &24.8\\
        (2) WMT$_{\rm 4L}$ &21.2 &14.7 &9.3 &13.2\\
        (2) $\rightarrow$ (1) &\bf 31.6 &\bf 19.5 &17.0 &\bf 26.4\\
        (1) $\rightarrow$ (2) &29.4 &18.0 &14.2 &21.4\\
        None &1.0 &0.7 &0.7 &2.8\\
        \bottomrule
        \multicolumn{5}{l}{\scriptsize \emph{$\rightarrow$: sequential training}}\\
        \end{tabular}
        \end{center}}
    \end{minipage}
    \begin{minipage}{0.23\linewidth}{\begin{center}
        \begin{tabular}{cc cccc}
        \toprule
        $\lambda_1$ &$\lambda_2$ &En &Zh &De &Fr \\
        \midrule
        \rowcolor{gray!10}
        0   &1      &\bf 29.9 &\bf 18.0 &\bf 17.1 &\bf 24.8\\
        0.1 &1      &28.6 &18.9 &15.0 &23.6\\
        1   &1      &27.5 &16.3 &12.6 &22.8\\
        1   &0.1    &24.9 &15.4 &13.1 &20.7\\
        1   &0      &23.1 &14.2 &10.3 &19.1\\
        \bottomrule
        \\
        \end{tabular}
    \end{center}}
    \end{minipage}
\end{table*}
%%%%%%%%%%%%%%%%%%%%%%%%%%%%%

\subsection{Quantitative Analysis}
In this section, we conduct a quantitative analysis to understand the contributions of each component in our approach.

\smallskip\noindent\myparagraph{Ablation study}
As shown in Table~\ref{tab:ablation_study}, removing our components significantly degrades the performance across different languages, which demonstrates the effectiveness of our proposed methods for zero-shot natural language generation (NLG).
In particular, both settings (b) and (c) show that when removing the cross-domain alignment, the model fails to accurately perform zero-shot NLG, which indicates the importance of bridging different domains for zero-shot NLG.
By comparing Full and setting (a), we can find that data corruption strategy can bring significant improvements.
It can be explained that the data corruption strategy in language reconstruction can encourage the model to efficiently bridge the domains and learn more robust latent representations.
We verify it in the following visualization.
Overall, the ablation study proves our arguments and the effectiveness of the proposed components.

\smallskip\noindent\myparagraph{Effect of the number of languages}
By comparing the performance of Full and settings (d-g) in Table~\ref{tab:ablation_study}, 
we can find that adding more languages successfully enables our ZeroNLG to not only deal with multiple language application scenarios within a single unified framework but also consistently boost the performance of each language application scenario across all metrics. The improved results show that the different knowledge from different language texts can be unified in our ZeroNLG to achieve a better language understanding, producing an overall improvement across all metrics regardless of the downstream language application scenarios. 

\smallskip\noindent\myparagraph{\YB{Effect of encoder choices}}
\YB{We compare different pre-trained multilingual encoders, including multilingual DistilBERT\cite{sanh2019distilbert} (\emph{mDistilBERT}), multilingual BERT\cite{devlin2018bert} (\emph{mBERT}), XLM-RoBERTa\cite{conneau2020XLM-R} (\emph{XLM-R}), and sentence BERT\cite{reimers2019Sentence-BERT} (\emph{sBERT}). All these models are base-size. As we can observe from Table~\ref{tab:more_ablation}, mBERT outperforms mDistilBERT due to a larger capacity. Surprisingly, XLM-R obtains inferior performance, possibly because it can not measure semantic textual similarity well\footnote{https://www.sbert.net/examples/training/multilingual.}. By contrast, sBERT — the model that has been pre-trained on a sentence similarity task, performs the best among all variants. Thus, our ZeroNLG is not limited to the default encoder choice (i.e., mDistilBERT) and can benefit from a more sophisticated encoder like sBERT.
}

\smallskip\noindent\myparagraph{\YB{Effect of decoder capacities}}
\YB
{The decoder of our ZeroNLG is shallow (i.e., $L=3$ layers) by default. We here ablate different $L$ in Table~\ref{tab:more_ablation}, where we can observe gradually boosted performance as $L$ increases. Therefore, our ZeroNLG can be improved by a larger decoder capacity.
It not only proves that more model parameters or stronger decoders can lead to further improvements, but also shows the potential of our ZeroNLG, which could be further improved by directly scaling up the model parameters.}

\smallskip\noindent\myparagraph{\YB{Effect of corpora}}
\YB{
Apart from the cross-lingual alignment corpus introduced in Section~\ref{sec:datasets} (abbreviated as CC3M$_{\rm 4L}$), we here consider translation data that focuses on news text for comparison and construct a corpus dubbed WMT$_{\rm 4L}$ of the same data scale as CC3M$_{\rm 4L}$. From Table~\ref{tab:more_ablation}, we can observe that although using WMT$_{\rm 4L}$ is not as competitive as CC3M$_{\rm 4L}$ due to substantial distribution shifts, it still achieves obvious performance gains compared with not using any corpora. 
Besides, switching the training orders of WMT$_{\rm 4L}$ and CC3M$_{\rm 4L}$ can produce encouraging results, proving the robustness of our model to training. More importantly, sequentially training on WMT$_{\rm 4L}$ and CC3M$_{\rm 4L}$ (i.e., ``(2) $\rightarrow$ (1)'') yields generally the best results. This demonstrates the importance of pre-training on corpora with gradually decreasing distribution shifts. 
}

\smallskip\noindent\myparagraph{\YB{Effect of loss designs}}
\YB{
In Table~\ref{tab:more_ablation}, we evaluate the effect of the InfoNCE loss ($\lambda_1$) and the MSE loss ($\lambda_2$) when performing cross-lingual alignment. We can observe that $\lambda_1 > 0$ suffers from performance degradation compared with $\lambda_1 = 0$. We speculate that this is because contrastive learning requires a large batch size to include sufficient (hard) negative samples\cite{chen2020simple}, which is hard to realize in our case due to resource restrictions. In contrast, $\{\lambda_1, \lambda_2\} = \{0, 1\}$ is compute-friendly and performs the best.
}

%%%%%%%%%%%%%%%%%%%%%%%%%%%%%
\begin{figure}[t]
\centering
\includegraphics[width=1\linewidth]{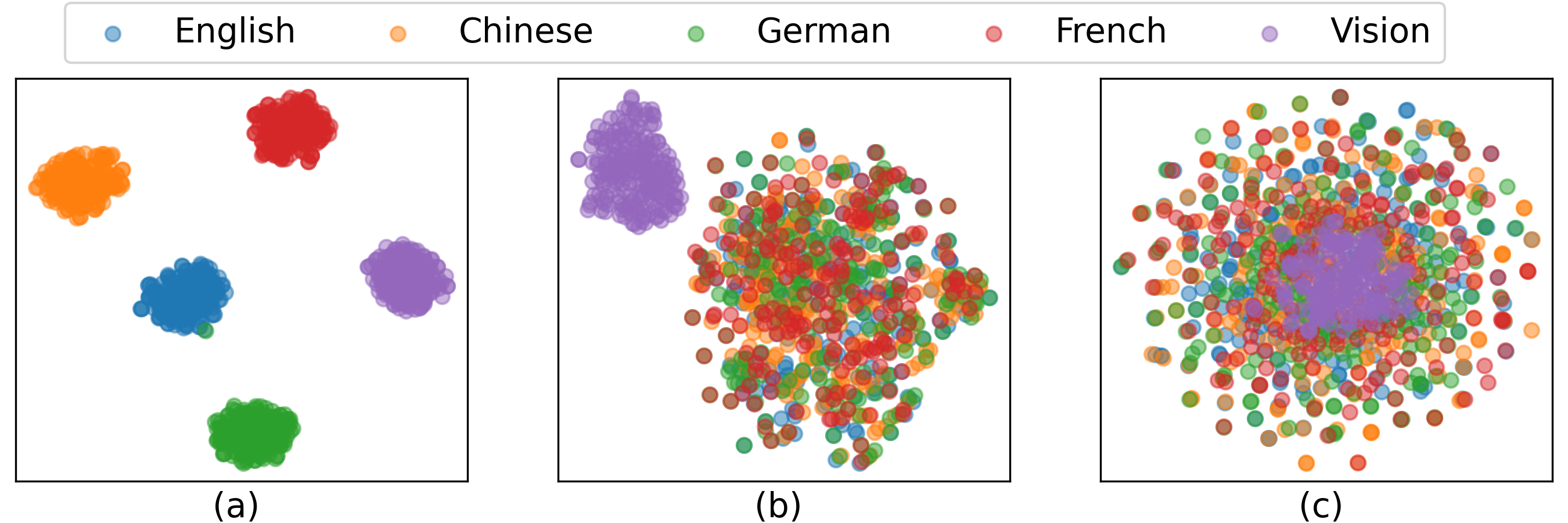}
\vspace{-15pt}
\caption{We show the t-SNE visualization \cite{tsne} of vision and multilingual embeddings. We plot the scatter diagrams with 200 samples for each modality and language. 
For comparison, we show the embeddings learned by (a) the Base model \YB{(i.e., without our CDA and DLR)}, (b) the Base model with \YB{CDA (Eq.~\ref{eq:formulation})}, and (c) our full model ZeroNLG.
}
\label{fig:ablation_tsne}
\end{figure}
%%%%%%%%%%%%%%%%%%%%%%%%%%%%%

%%%%%%%%%%%%%%%%%%%%%%%%%%%%%
\begin{figure*}[t]
\centering
\includegraphics[width=1\linewidth]{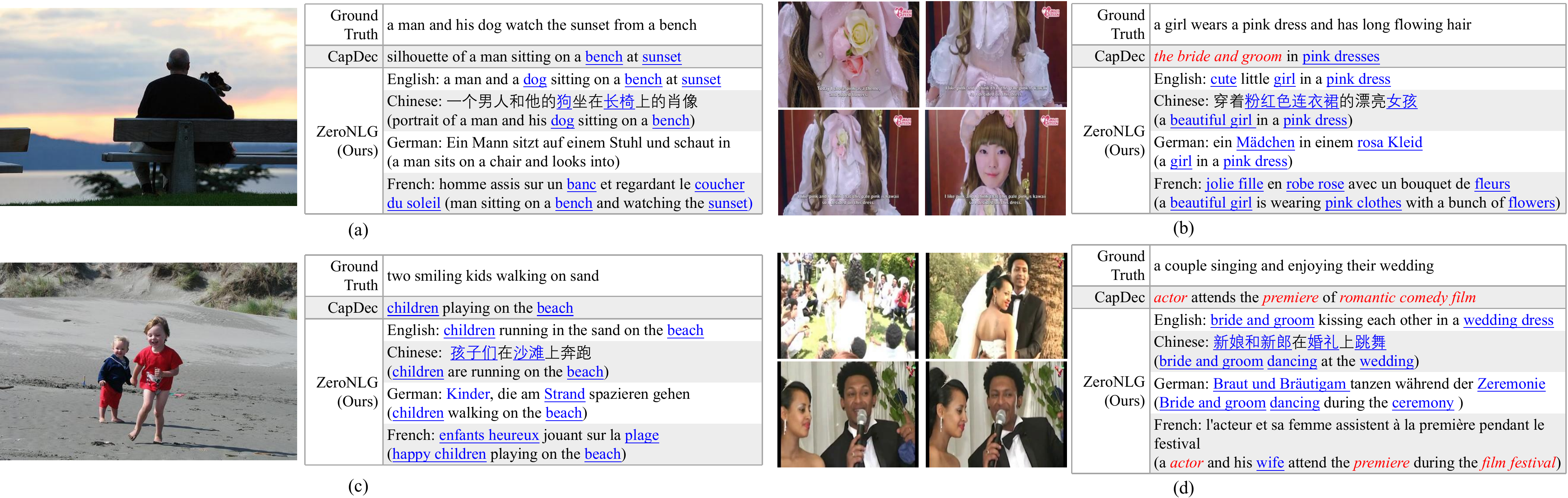}
\vspace{-15pt}
\caption{The examples of visual captions generated by the state-of-the-art zero-shot model \cite{nukrai2022text} and our ZeroNLG model for different languages, i.e., English, Chinese, German, and French, under the zero-shot setting.
For better understanding, we add English translations below the non-English captions in brackets. 
We highlight \textcolor{blue}{\underline{{accurate}}} keywords and \textcolor{red}{\emph{wrong}} details. 
As we can see,
% Although our ZeroNLG is not trained with vision-text data, 
ZeroNLG can generate accurate and vivid descriptions with more visual details across languages.
}
\label{fig:qualitative_examples}
\end{figure*}
%%%%%%%%%%%%%%%%%%%%%%%%%%%%%

\subsection{Visualization}
To better understand our method, in Figure~\ref{fig:ablation_tsne}, we adopt t-SNE \cite{tsne} to visualize \YB{vision and multilingual embeddings}.
For comparison, \YB{we consider (a) the Base model (i.e., without our proposed CDA and DLR), (b) the Base model with CDA, and (c) our ZeroNLG.}
As we can see, for the Base model, although the embeddings belonging to the same domain are well-clustered, there is a clear gap between different domains.
\YB{With CDA, texts in different languages are well-aligned with each other whereas vision and text domains still exhibit disparities, possibly because contrastive learning converges to local minima\cite{shi2023towards}.}
\YB{Finally, by considering data corruption during DLR, our ZeroNLG can align and bridge different domains across modalities and languages well.}

\subsection{Qualitative Analysis}
In this section, we conduct a qualitative analysis in Figure~\ref{fig:qualitative_examples} to intuitively understand our proposed ZeroNLG.

\smallskip\noindent\myparagraph{Case study}
In Figure~\ref{fig:qualitative_examples}, we give four examples to compare our proposed ZeroNLG with the state-of-the-art (SoTA) zero-shot model CapDec \cite{nukrai2022text}. Both models are trained on the same pre-training corpus and have no access to downstream vision-text data. 
As we can observe, under the zero-shot setting, the SoTA can not well describe the vision content and generates several wrong descriptions (Red-colored text), failing to depict some important visual objects, e.g., ``\textit{dog}'' in (a), ``\textit{girl}'' and ``\textit{flowers}'' in (b), and ``\textit{bride and groom}'' in (d), while our approach can generate fluent and ``believable'' captions containing these important objects.
More encouragingly, the captions generated by our ZeroNLG are well supported by accurate visual details,  e.g., ``\textit{beautiful}'' and ``\textit{cute}'' in (b), ``\textit{happy}'' in (c), and ``\textit{wedding}'' in (d).
Besides, considering that there are no public human-annotated datasets available for the video to German/French tasks, the capacity of our ZeroNLG in generating reasonable German and French captions is encouraging. 
Overall, our approach can generate high-quality and desirable outputs for different languages within a unified framework. 
It further proves our arguments and the effectiveness of our proposed approach.

\smallskip\noindent\myparagraph{Error analysis}
We further perform the error analysis to analyze our ZeroNLG.
As we can see, our model suffers from several common drawbacks: i) generating repeated or incomplete sentences, e.g., \textit{``in the sand on the bench''} and the German sentence in (a); and ii) misunderstanding objects and scenes in some cases, e.g., the French sentence in (d).
They can be attributed to the lack of detailed visual relationships and accurate visual information.
We may alleviate these drawbacks by introducing a visual object extractor and a scene graph, where the former predicts a set of visual objects and the latter models the relationships between objects.
Both of them are widely used in previous zero-shot methods \cite{Feng2019Unsupervised_ic,gu2019unpaired,Yang2020USGAE}.
However, it is unlikely to be avoided completely, as these drawbacks are common in natural language generation \cite{See2017Get}.

\section{Conclusions and Future Works}
In this work, we make the first attempt to achieve zero-shot multimodal and multilingual natural language generation in a unified framework. 
To this end, we propose the intuitive ZeroNLG approach, which first exploits English-centric data to align and bridge different domains across modalities and languages, and then auto-encodes languages to learn to perform zero-shot NLG.
Our experiments demonstrate that, without any available downstream data for training, ZeroNLG can produce desirable outputs given various forms of input data, i.e. images, video, and text. 
Extensive investigation of performance with twelve NLG tasks, including image captioning, video captioning, and neural machine translation, demonstrate the effectiveness of our approach, and where we conclude that ZeroNLG significantly outperforms previous state-of-the-art zero-shot methods within a single framework.

\textbf{Future works:} Substantial research avenues exist in i) further aligning with various types of images, such as artwork and 3D renderings, to perform the image-to-image translation and text-to-image translation (i.e., image generator) within a unified framework;
ii) further boosting performance by learning and unifying knowledge from various different languages (and images).
We note in passing that our proposed method can be improved by incorporating more languages, which has been shown in Table~\ref{tab:ablation_study}.  We emphasize the applicability of this method in improving the manner in which we offer data-to-data mappings for under-represented languages and communities, which is an area of particular future opportunity in promoting ``fair AI''.

% or


% \appendices

% % you can choose not to have a title for an appendix
% % if you want by leaving the argument blank
% \section{}

% % use section* for acknowledgment
% \ifCLASSOPTIONcompsoc
%   % The Computer Society usually uses the plural form
%   \section*{Acknowledgments}
% \else
%   % regular IEEE prefers the singular form
% \fi


\ifCLASSOPTIONcaptionsoff
  \newpage
\fi



%\bibliographystyle{IEEEtran}

\bibliographystyle{IEEEtran}
\bibliography{ref,ref2}


\begin{IEEEbiography}[{\includegraphics[width=1in,height=1.25in,clip,keepaspectratio]{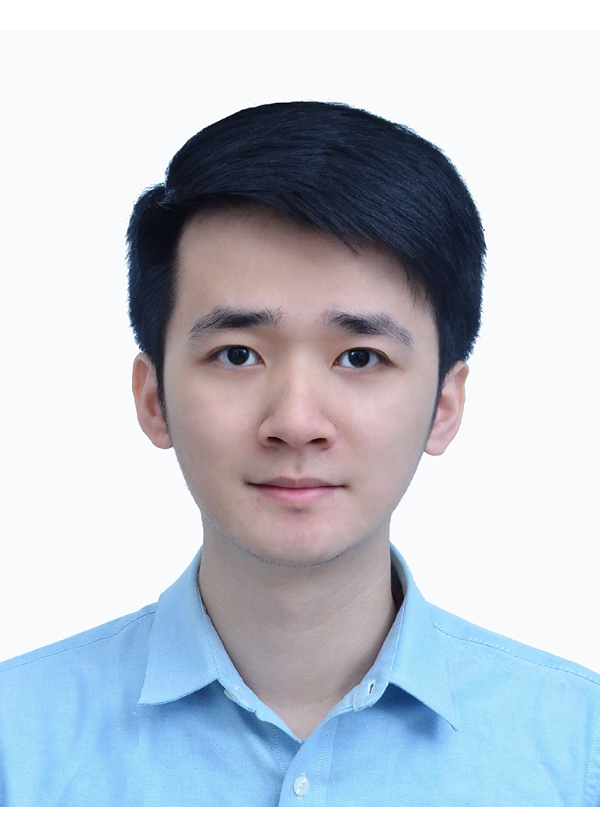}}]{Bang Yang} received the B.E. degree from Sun Yat-sen University in 2018 and the M.S. degree from Peking University in 2021, where he is currently pursuing the Ph.D. degree. He is engaged in a joint training program with the Pengcheng Laboratory. His research interests include multimodal learning and AI in healthcare.
\end{IEEEbiography}

\begin{IEEEbiography}[{\includegraphics[width=1in,height=1.25in,clip,keepaspectratio]{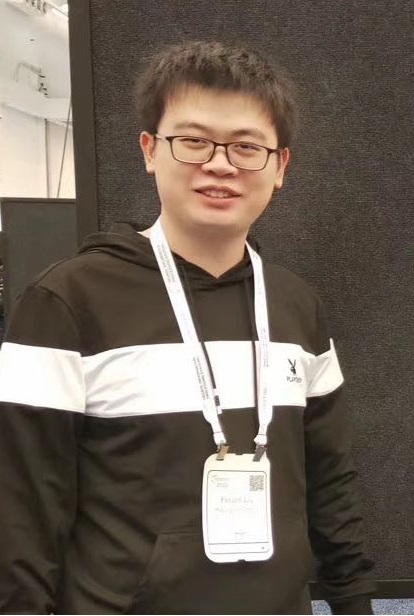}}]{Fenglin Liu} 
% \begin{IEEEbiography}{Fenglin Liu}
is a PhD student at the University of Oxford. His research interests include Vision-and-Language Processing, Machine Learning, and their applications to healthcare.
He has published papers at premier journals and conferences, e.g., TPAMI, NeurIPS, CVPR, ACL, EMNLP, NAACL. He has served as a senior program committee member for IJCAI and was awarded as the Distinguished/Outstanding Reviewer of CVPR, AAAI, and IJCAI.
\end{IEEEbiography}

\begin{IEEEbiography}[{\includegraphics[width=1in,height=1.25in,clip,keepaspectratio]{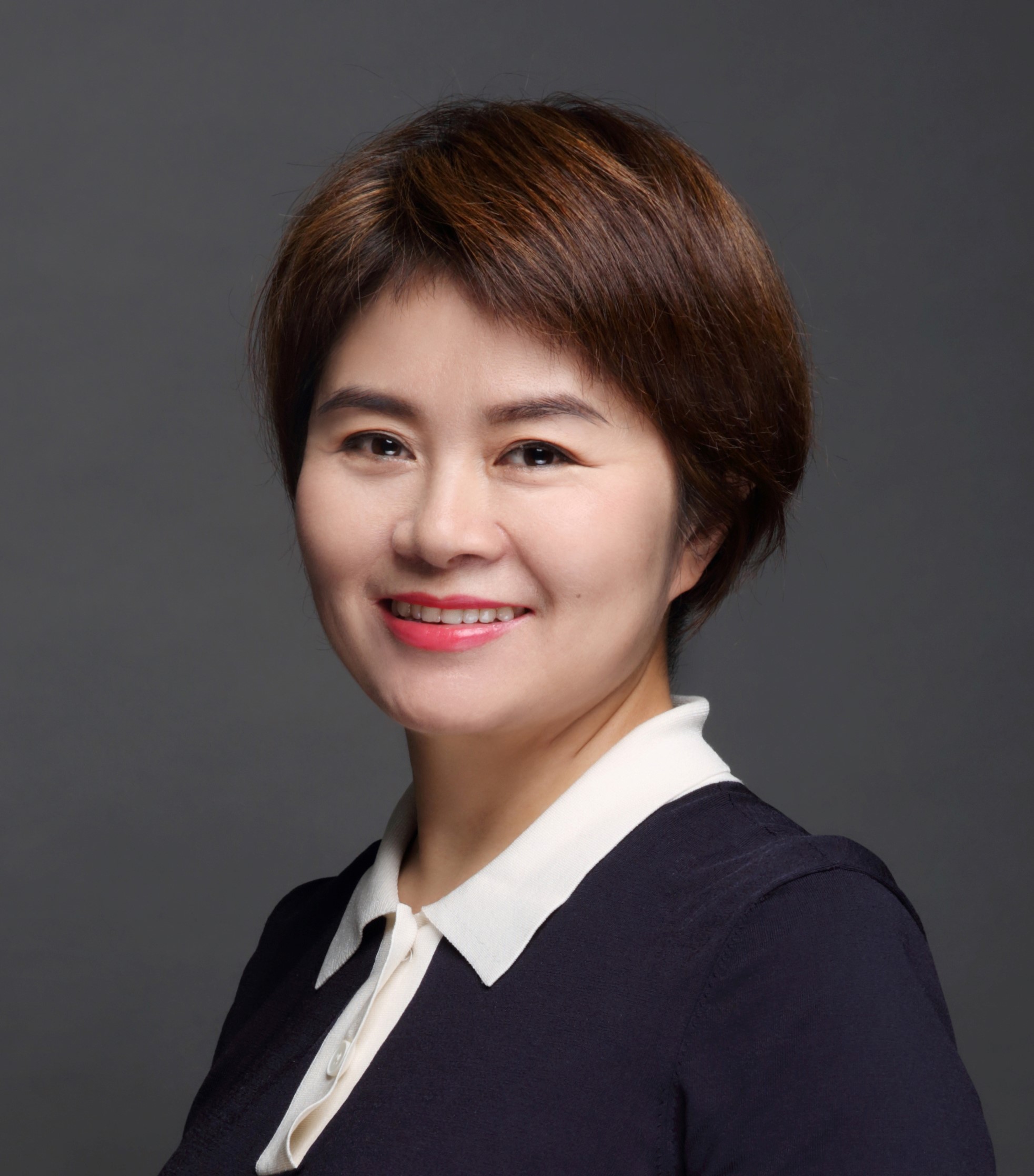}}]{Yuexian Zou} (Senior Member, IEEE) 
% received her B.Sc. degree from the University of Electronic Science and Technology in 1985 and Ph.D. degree from the University of Hong Kong in 2001, respectively. She 
is currently a Full Professor with Peking University and the Director of the Advanced Data and Signal Processing Laboratory in Peking University and serves as the Deputy Director of Shenzhen Association of Artificial Intelligence (SAAI).
She was a recipient of the award Leading Figure for Science and Technology by Shenzhen Municipal Government in 2009 and now is the adjunct professor in Pengcheng Laboratory. 
% Since 2010, she has been actively involved in teaching and research on machine learning and its applications in video and audio analysis. 
She conducted more than 20 research projects including NSFC and 863 projects. She has published more than 280 academic papers in famous journals and flagship conferences, and issued nine invention patents.
% She conducts several courses for graduate students, such as machine learning and pattern recognition, digital signal processing, and array signal processing. 
Her research interests are mainly in machine learning and scene understanding.
\end{IEEEbiography}

\begin{IEEEbiography}[{\includegraphics[width=1in,height=1.25in,clip,keepaspectratio]{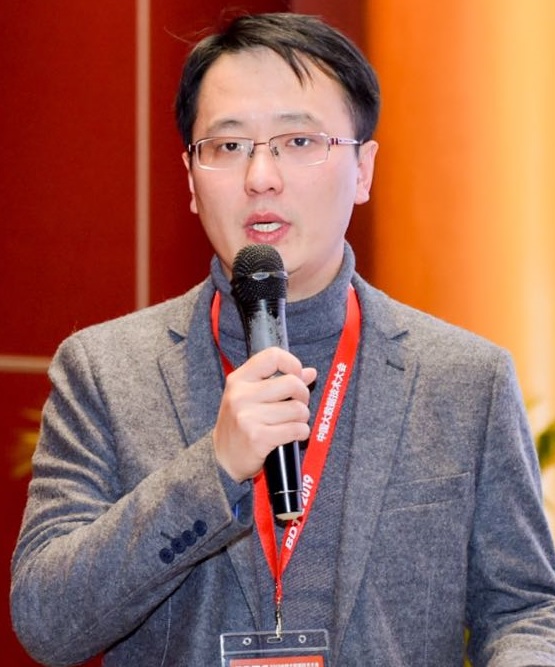}}]{Xian Wu} is now a Principal Researcher in Tencent. Before joining Tencent, he worked as a Senior Scientist Manager and a Staff Researcher in Microsoft and IBM Research. Xian Wu received his PhD degree from Shanghai Jiao Tong University. His research interests include Medical AI, Natural Language Processing and Multi-Modal modeling. Xian Wu has published papers in CVPR, NeurIPS, ACL, WWW, AAAI, IJCAI etc. He also served as PC member of TKDE, TKDD, TOIS, TIST, CVPR, ICCV, AAAI etc.
\end{IEEEbiography}

\begin{IEEEbiography}[{\includegraphics[width=1in,height=1.25in,clip,keepaspectratio]{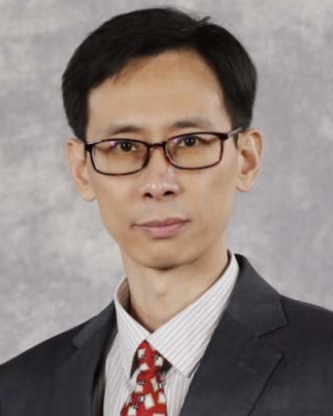}}]{Yaowei Wang} (Member, IEEE) received the Ph.D. degree in computer science from the Graduate University of Chinese Academy of Sciences in 2005. He is currently an Associate Professor with the Pengcheng Laboratory, Shenzhen, China. He was a Professor at the National Engineering Laboratory for Video Technology Shenzhen (NELVT), Peking University Shenzhen Graduate School, in 2019. From 2014 to 2015, he worked as an Academic Visitor at the Vision Laboratory, Queen Mary University of London. He worked at the Department of Electronics Engineering, Beijing Institute of Technology, from 2005 to 2019. His research interests include machine learning, multimedia content analysis, and understanding. He is the author or coauthor of over 70 refereed journals and conference papers. He was a recipient of the second prize of the National Technology Invention in 2017 and the first prize of the CIE Technology Invention in 2015. His team was ranked as one of the best performers in the TRECVID CCD/SED tasks from 2009 to 2012 and PETS in 2012. He is a member of CIE, CCF, and CSIG.
\end{IEEEbiography}

\begin{IEEEbiography} 
[{\includegraphics[width=1in,height=1.25in,clip,keepaspectratio]{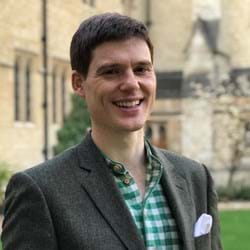}}]{David A. Clifton} is the Royal Academy of Engineering Chair of Clinical Machine Learning at the University of Oxford, and OCC Fellow in AI \& Machine Learning at Reuben College, Oxford.  He was the first AI scientist to be appointed to an NIHR Research Professorship, which is the UK medical research community's ``flagship Chair programme''.  He is a Fellow of the Alan Turing Institute, Research Fellow of the Royal Academy of Engineering, Visiting Chair in AI for Healthcare at the University of Manchester, and a Fellow of Fudan University, China.
He studied Information Engineering at Oxford's Department of Engineering Science, supervised by Prof. Lionel Tarassenko CBE, Chair of Electrical Engineering.  His research focuses on the development of machine learning for tracking the health of complex systems. His previous research resulted in patented systems for jet-engine health monitoring, used with the engines of the Airbus A380, the Boeing 787 ``Dreamliner'', and the Eurofighter Typhoon. Since graduating from his DPhil in 2009, he has focused mostly on the development of AI-based methods for healthcare. Patents arising from this collaborative research have been commercialised via university spin-out companies OBS Medical, Oxehealth, and Sensyne Health, in addition to collaboration with multinational industrial bodies.  He was awarded a Grand Challenge award from the UK Engineering and Physical Sciences Research Council, which is an EPSRC Fellowship that provides long-term strategic support for ``future leaders in healthcare''.  His research has been awarded over 35 academic prizes; in 2018, he was joint winner of the inaugural ``Vice-Chancellor's Innovation Prize'', which identifies the best interdisciplinary research across the entirety of the University of Oxford.
\end{IEEEbiography}

%\newpage


%\vfill


\end{document}